# Foreign-Object Detection in High-Voltage Transmission Line Based on Improved YOLOv8m

**Zhenyue Wang [1,2], Guowu Yuan [1,3,\*], Hao Zhou [1,3], Yi Ma [2] and Yutang Ma [2]**

[1] School of Information Science and Engineering, Yunnan University, Kunming 650504, China; zywang@mail.ynu.edu.cn (Z.W.); zhouhao@ynu.edu.cn (H.Z.)

[2] Electric Power Research Institute, Yunnan Power Grid Co., Ltd., Kunming 650214, China; mayi@dlyjy.yn.csg.cn (Y.M.); mayutang@dlyjy.yn.csg.cn (Y.M.)

[3] Yunnan Key Laboratory of Intelligent Systems and Computing, Kunming 650504, China

\* Correspondence: gwyuan@ynu.edu.cn

**Abstract:** The safe operation of high-voltage transmission lines ensures the power grid's security. Various foreign objects attached to the transmission lines, such as balloons, kites and nesting birds, can significantly affect the safe and stable operation of high-voltage transmission lines. With the advancement of computer vision technology, periodic automatic inspection of foreign objects is efficient and necessary. Existing detection methods have low accuracy because foreign objects attached to the transmission lines are complex, including occlusions, diverse object types, significant scale variations, and complex backgrounds. In response to the practical needs of the Yunnan Branch of China Southern Power Grid Co., Ltd., this paper proposes an improved YOLOv8m-based model for detecting foreign objects on transmission lines. Experiments are conducted on a dataset collected from Yunnan Power Grid. The proposed model enhances the original YOLOv8m by incorporating a Global Attention Module (GAM) into the backbone to focus on occluded foreign objects, replacing the SPPF module with the SPPCSPC module to augment the model's multiscale feature extraction capability, and introducing the Focal-EIoU loss function to address the issue of high- and low-quality sample imbalances. These improvements accelerate model convergence and enhance detection accuracy. The experimental results demonstrate that our proposed model achieves a 2.7% increase in mAP_0.5, a 4% increase in mAP_0.5:0.95, and a 6% increase in recall.

**Keywords:** transmission lines; foreign-object detection; YOLO; GAM; SPPCSPC; Focal-EIoU

## 1. Introduction

High-voltage transmission lines play a crucial role in ensuring the safety and stability of electricity transmission [1]. With the rapid development of the Chinese economy, China's power generation accounted for approximately 30.34% of the world's total in 2022, making its high-voltage transmission line network the largest globally [2]. These transmission lines traverse diverse and dynamic environments, often passing through residential and commercial areas, making them susceptible to the attachment of various foreign objects such as balloons, kites, fabrics and plastic waste [3]. Additionally, in natural settings like forests, birds nesting and twigs can challenge the safety of the transmission lines [4]. Suppose foreign objects on the transmission lines are not promptly detected and cleared: in that case, they can lead to accidents such as single-phase grounding and inter-phase short circuits, severely affecting the regular operation of the transmission lines, jeopardizing human safety, and resulting in casualties like accidental electrocution and fatalities [5].

Early detection of foreign objects on transmission lines relied predominantly on manual inspections, a method characterized by high labor intensity and inefficiency. With the development of computer vision technology and the widespread application of unmanned aerial vehicles (UAVs), UAVs can capture the transmission lines' images. Then, the images can be analyzed using computer vision techniques to automatically identify anomalies, presenting a significant advancement in line inspection methodologies [6,7].

Currently, target detection research on transmission lines has mainly focused on target detection of principal components, foreign-object detection, smoke detection, insulator defect detection, etc. Among them, foreign-object detection mainly focuses on four types of foreign objects, namely balloons, bird nests, kites, and plastic trash. However, in the plateau mountainous environment of Yunnan Province, China, slender twigs and large birds also seriously interfere with the regular operation of high-voltage transmission lines. Therefore, slender twigs and birds are added to our detection database. The existing methods cannot effectively solve the problems faced by the Yunnan Branch of China Southern Power Grid Co., Ltd., so our research has specific value.

This paper's contributions are as follows:

(1) Based on the data provided by the Yunnan Power Science Research Institute in the Yunnan Branch of China Southern Power Grid Co., Ltd., a foreign object dataset of power transmission lines was established. This dataset includes six types of objects: trash, twigs, nests, kites, birds, and balloons. In this database, foreign objects are occluded, with multiple types, large-scale changes, and complex backgrounds. The detection accuracy of existing methods is not high.

(2) We propose a foreign-object-detection model for high-voltage transmission lines based on improved YOLOv8m. The experimental results show that our model has achieved high accuracy and can meet the needs of practical applications.

## 2. Related Work

### 2.1. Object-Detection Model

In 2014, Girshick et al. [8] used the "region proposal + convolutional neural network" to replace the traditional method of "sliding window + manually designed features" in object detection, and designed the R-CNN framework, which made a huge breakthrough in object-detection technology. In 2015, Girshick et al. [9] proposed the Fast RCNN algorithm, which further improved RCNN and innovatively proposed multi task loss. They also trained classifiers and bounding box regressors, achieving end-to-end training in the detection stage, greatly improving accuracy and speed. Ren et al. [10] proposed the Faster RCNN algorithm shortly after, introducing an RPN network, making candidate box generation almost cost free. In 2016, Redmon et al. [11] proposed the YOLO algorithm, which is the first single-stage object-detection algorithm in the era of deep learning, and its speed is very fast. This network divides the image into grids and predicts the bounding box and classification probability of each grid area. A single neural network can obtain results from the complete image through one operation, which is conducive to end-to-end optimization of detection performance. In the same year, combining the anchor mechanism of RCNN and the regression idea of YOLO, Liu et al. [12] proposed the SSD algorithm, introducing multi-scale detection methods, and performing detection on feature maps extracted at each scale. Lin et al.[13] proposed RetinaNet in 2017 to investigate the accuracy of single-stage detection methods that lag behind two-stage detection methods. They believe that the imbalance of categories during the training process leads to a disadvantage in the accuracy of single-stage methods. Therefore, they propose Focal Loss to replace traditional cross entropy, improve the weight of background samples, and make the model more inclined towards target samples that are difficult to detect during the training process.

In 2017 and 2018, Redmon et al. launched subsequent versions of YOLO: YOLOv2 [14] and YOLOv3 [15]. The YOLOv2 algorithm introduces BN (batch normalization), multi-scale training, anchor box mechanism, and fine-grained features to improve the YOLOv1 algorithm. On the basis of YOLOv2, the YOLOv3 algorithm adopts better backbone network, multi-scale prediction, and nine anchor boxes for detection, which improves the accuracy of the detection algorithm while ensuring real-time performance. In 2020, Bochkovskiy A et al. [16] proposed the YOLOv4 model. In the same year, Glenn et al. [17] proposed YOLOv5, which greatly reduced the number of parameters in the YOLO model, but its accuracy was comparable to YOLOv4. In 2021, Ge Zheng et al. [18] proposed the YOLOX model. In the same year, Wang et al. [19] proposed YOLOR. In 2022, Meituan [20] proposed YOLOv6, which was inspired by the design concept of hardware-aware neural networks. Based on the RepVGG style [21], it designed the reconfigurable and more efficient backbone networks, EfficientRep Backbone and Rep-PAN Neck; it optimized and designed a more concise and effective Efficient Decoupled Head, which further reduces the additional delay overhead caused by general decoupling heads while maintaining accuracy; and the anchor-free mode was adopted, supplemented by the SimOTA [22] label allocation strategy and SIoU [23] boundary box regression loss, to further improve detection accuracy. In the same year, Wang, C, et al. [24] proposed YOLOv7, proposing different network models for different devices. The concept of a gradient propagation path was used to analyze the reparameterization strategies applicable to different layers of networks, and a planned model structure reparameterization was proposed; the author proposes a new label allocation method called the Coarse to Fine (Coarse to Fine) guided label allocation strategy, which allows Aux Head to generate more grids as positive samples and reduce information loss. In terms of model design architecture, E-ELAN was proposed, which enhances the network feature's learning ability through amplification, confusion, and merging without changing the original gradient path of ELAN [25]. RepVGG style can improve performance through multiple branches during the training process, and inference speed can be accelerated through structural reparameterization. In 2023, Ultratics [26] released YOLOv8, which may have referenced the design concept of YOLOv7 ELAN in its backbone network and Neck section. YOLOv5's C3 structure was replaced with a C2f structure with richer gradient flow, and different channel numbers were adjusted for different scale models, significantly improving model performance; the Head section has been replaced with the current mainstream decoupling head structure, separating classification and detection heads, and also replaced with Anchor Free from Anchor Based. In terms of loss calculation, the TaskAlignedAssigner positive sample allocation strategy was adopted, and the Distribution Focal Loss was introduced. The data augmentation part of the training introduces the last 10 epochs in YOLOX to turn off Mosaic augmentation, which can effectively improve accuracy.

YOLOv8m is one of the latest models in object detection, exhibiting advantages in detection accuracy and speed. Because the foreign-object-detection model for high-voltage transmission lines will be integrated into the hardware of the drone platform in the later stage, we are considering using the YoloV8m lightweight model and modifying it to make it more suitable for our application.

### 2.2. Research on Target Detection for Transmission Lines

In the early stages of employing computer vision for detecting foreign objects on transmission lines, traditional digital-image processing techniques were often used, employing non-neural network-based methods to extract defect features from transmission line images [27]. For instance, Cai et al. [28] employed contour detection of the transmission conductor's shape to determine the presence of foreign objects. Jin et al. [29] proposed an Otsu adaptive threshold segmentation algorithm, improved with morphology, to remove background noise, and applied a gradient method to identify power line edge positions, utilizing the Hough transform to analyze the number of lines to achieve foreign object recognition. Wang et al. [30] introduced a detection method for foreign objects on transmission lines based on line structure perception, segmenting and analyzing the grayscale and line width of acquired line segments in horizontal and vertical directions, to identify foreign objects on the transmission lines within complex backgrounds. However, the rich and diverse details in images of foreign objects on transmission lines make it challenging to describe target features using manually designed single features comprehensively. These methods exhibit relatively low accuracy in identifying foreign objects and struggle to differentiate between anomalies.

With the rapid advancement of deep learning, methodologies for detecting foreign objects on power transmission lines based on deep learning have been proposed. Shi et al. [31] initially employed the Selective Search method to extract candidate regions from pole images. Subsequently, training was conducted using the CaffeNet network model, adjusting and optimizing samples and network parameters through pre-training and re-training, resulting in a detection accuracy of 92.46% for nests. Zhu et al. [32] processed images using the scale histogram-matching method and employed the DFB-NN model, which effectively integrates low-resolution information and high-semantic information, to detect various objects, including vehicle-mounted cranes, tower cranes, forklifts, winding foreign objects, and wildfires, achieving an accuracy of 88.1%. Yang et al. [33] utilized the DenseNet network to replace the penultimate layer network in YOLO v3, achieving average detection accuracies of 94.7% for kites, nests, and trash. Yu et al. [34] used Otsu adaptive threshold segmentation, morphological processing and other methods to extract regions of interest. Then, DenseNet201 was used to extract the features of the regions of interest. Finally, the ECOC-SVM model was trained, and the average accuracy of balloons, kites, plastics and nests reached 93.3%. Yu et al. [35] employed a concatenated fusion strategy to merge features extracted by different networks. They utilized a random-forest classification model to achieve a detection accuracy of 95.8% for balloons, kites, nests, and plastics. Zou et al. [36] replaced the FPN structure of YOLOv5 with the BiFPN structure, integrated the CA attention mechanism into the CSP2_X module, and added a small-object detection layer, resulting in an average detection accuracy of 98.3% for nests, plastics, and kites. Qiu et al. [37] proposed the YOLOv4-EDAM model, utilizing the DnCNN image denoising network to reduce image noise and employing SENet-MobileNetV2 to extract features. Additionally, the model enhanced the network's feature extraction capability using CBAM-SPP and CBAM-PANet, and utilized NMS to eliminate overlapping boxes, achieving a detection accuracy of 96.71% for nests, kites, balloons and trash. Wu et al. [38] used the ASPP module in YOLOX, incorporated the CBAM attention mechanism, employed the GIOU loss function, and achieved a detection accuracy of 86.57% for nests, balloons, trash and kites. Yu et al. [39] introduced SPD convolution into the YOLOv7 model, optimizing hyperparameters to achieve a detection accuracy of 92.2% for cranes, excavators, bulldozers, tower cranes, trucks and nests. Zhang et al. [40] made improvements based on the YOLOv4 model, generating anchor boxes suitable for this dataset using k-means clustering, replacing the network's SPP module with an SPPF module, and substituting the SiLU activation function for the Leaky ReLU activation function, achieving an average detection accuracy of 97.57% for balloons, nests, kites, and trash. Tang et al. [41] combined Transformer V2 with YOLOX, utilizing the STCSP feature extraction layer in the backbone network and employing the Hybrid Spatial Pyramid Pooling (HSPP) module and the RepVGGBlock. This approach achieved a detection accuracy of 96.7% for nests, kites, and balloons. In research on transmission line components and environment, Chen et al. [42] improved a YOLOv5s model for key-component detection of power transmission lines. Cheng et al. [43] proposed a detection algorithm called P2E-YOLOv5 to detect a small target. Chen et al. [44] proposed an optimized YOLOv7-tiny model for smoke detection in power transmission lines.

The existing foreign-object detection mainly focuses on four types of foreign objects, namely balloons, bird nests, kites, and plastic trash. However, in the plateau mountainous environment of Yunnan Province, China, slender twigs and large birds also seriously interfere with the regular operation of high-voltage transmission lines. Therefore, slender twigs and birds are added to our detection database. The existing methods cannot effectively solve the problems faced by the Yunnan Branch of China Southern Power Grid Co., Ltd.

## 3. Datasets

The datasets of foreign objects on power transmission lines utilized in this study were provided by the Electric Power Science Research Institute in Yunnan Branch of China Southern Power Grid Co., Ltd. The image data was collected using various drones, resulting in significant variations in pixel dimensions, ranging from 640 × 640, 856 × 486, 1280 × 720 to 2738 × 2270. The sample images of foreign objects on power transmission lines were categorized into six classes: trash, twig, nest, kite, bird and balloon. Figure 1a represents trash, Figure 1b depicts twig, Figure 1c displays nest, Figure 1d exhibits kite, Figure 1e portrays bird, and Figure 1f illustrates balloon. It is essential to note that the images shown in Figure 1 are not original. Because the foreign objects are small, proportionally, in some original images, we extracted the partial images around foreign objects for a more explicit representation.

Due to detailed latitude and longitude information in some sample images, we have covered this information with black blocks according to data security requirements.

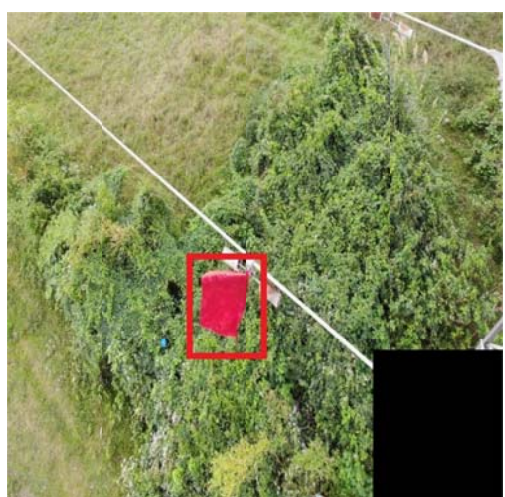
(a)

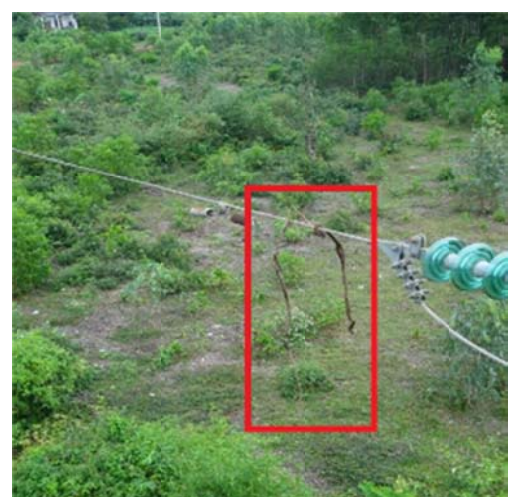
(b)

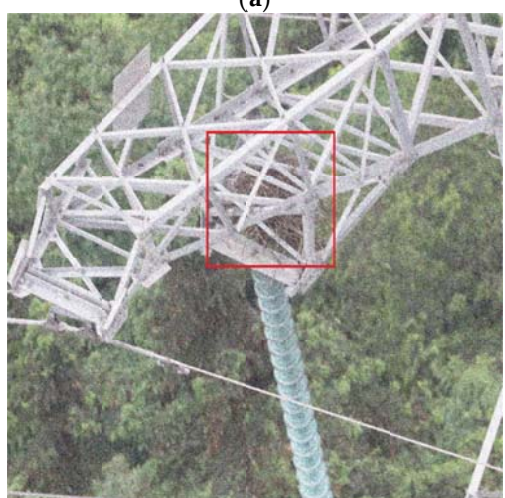
(c)

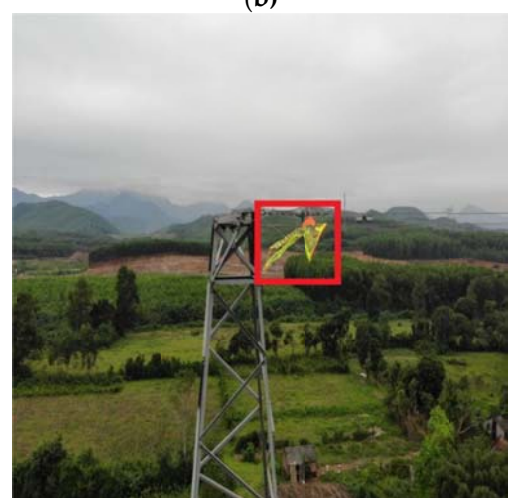
(d)

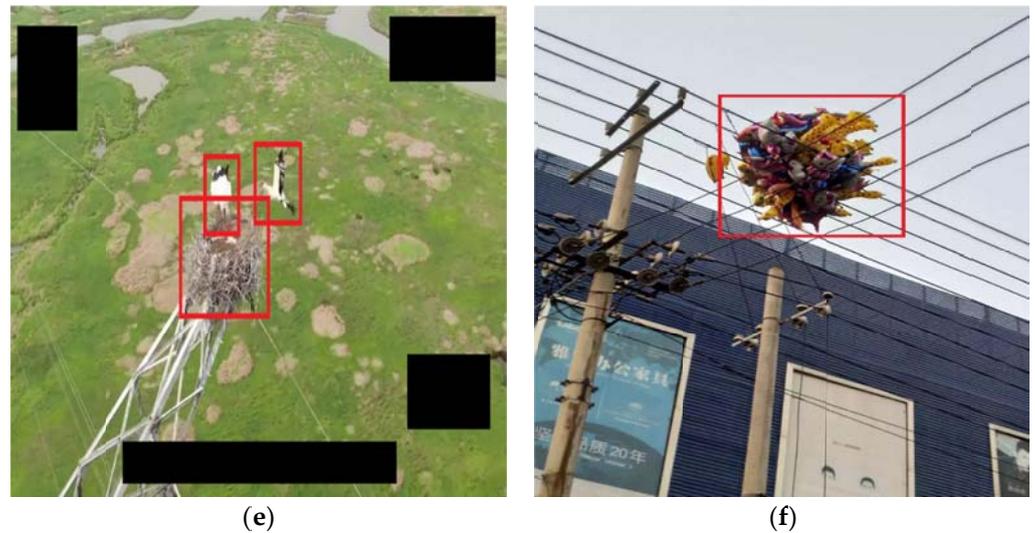

|(**e**)|(**f**)|

**Figure 1.** Collection of Foreign Object Image Datasets for Transmission Lines. (**a**) trash. (**b**) twig. (**c**) nest. (**d**) kite. (**e**) bird. (**f**) balloon.

The overall images for each category are presented in Figure 1. It is evident from Figure 1 that, due to varying shooting angles, the images of foreign objects on power transmission lines exhibit challenges such as occlusions, diverse categories of foreign objects, considerable variations in object scales, and complex backgrounds. These factors significantly augment the difficulty in detecting foreign objects on power transmission lines.

Given the infrequent occurrence of foreign objects on power transmission lines, the samples obtained are limited. In the original dataset, the respective counts of actual labels for trash, twig, nest, kite, bird and balloon are 667, 64, 2508, 240, 1244, and 356, totaling 3774 images in the dataset. These samples are insufficient for training neural networks, necessitating dataset augmentation.

As depicted in Figure 2, below, we augmented the datasets through rotations, variations in brightness, salt-and-pepper noise, and occlusions, resulting in an expanded dataset comprising 11,323 images. Following augmentation, the counts of actual labels for trash, twig, nest, kite, bird, and balloon are 2024, 1299, 4017, 2133, 2489, and 1811, respectively. The quantities per category are detailed in Table 1.

In subsequent experiments, to assess the effectiveness of the improved model, we randomly divided both the original and augmented dataset into training set, validation set, and test sets in a 6:2:2 ratio, for experimentation with the proposed model.

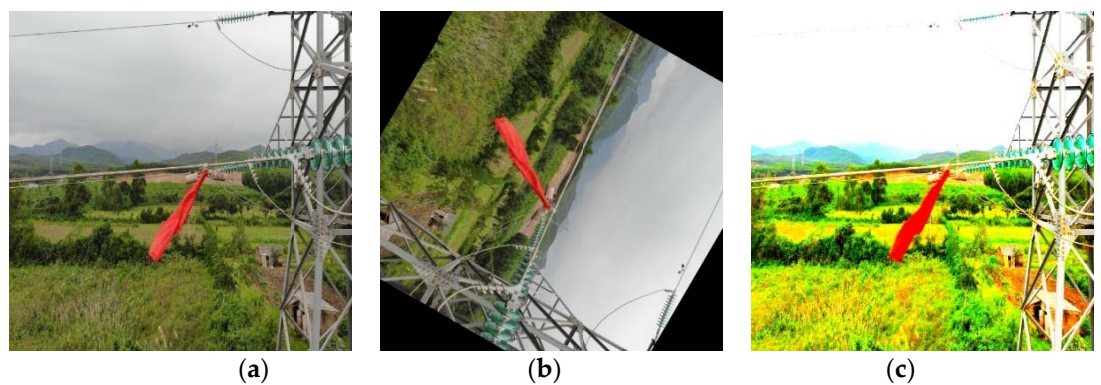

|(**a**)|(**b**)|(**c**)|

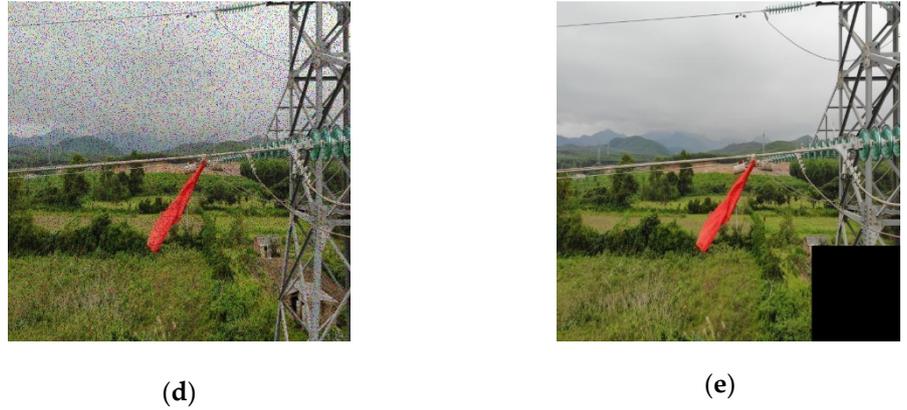

(**d**)           (**e**)

**Figure 2.** Example of data augmentation. (a) original. (b) rotations. (c) variations in brightness. (d) salt-and-pepper noise. (e) occlusions.

**Table 1.** Total Amount of Each Type of Data.

| Type | Trash | Twig | Nest | Kite | Bird | Balloon |
|------|-------|------|------|------|------|---------|
| Original | 667 | 64 | 2508 | 240 | 1244 | 356 |
| Data Augmentation | 2024 | 1299 | 4017 | 2133 | 2489 | 1811 |

## 4. Proposed Method

Addressing the challenges posed by occlusions, diverse object categories, significant variations in object scales, and complex backgrounds in detecting foreign objects on power transmission lines, this paper proposes an improved model for foreign-object detection based on the YOLOv8m framework. A GAM attention mechanism is incorporated into the backbone to enhance the model's focus on the foreign object, to mitigate their occlusion. In response to the challenge of multiple object categories and intricate backgrounds, the SPPF module is replaced with the SPPCSPC module to augment the model's multi-scale feature-extraction capability. A Focal-EIOU loss function is introduced to address the imbalance in high- and low-quality samples, thereby expediting model convergence and elevating detection precision. These enhancements are elaborated sequentially in Sections 3.2 to 3.4.

### 4.1. Enhanced YOLOv8m Model

The improved YOLOv8m model is illustrated in Figure 3, and the enhancements are outlined within the red boxes, elucidated as follows.

In Figure 3, the GAM attention mechanism is introduced into the backbone, a feature detailed in Section 4.2. The SPPF module is replaced by the SPPCSPC module, a modification discussed in Section 4.3. The integration of the Focal-EIoU loss function is presented in Section 4.4.

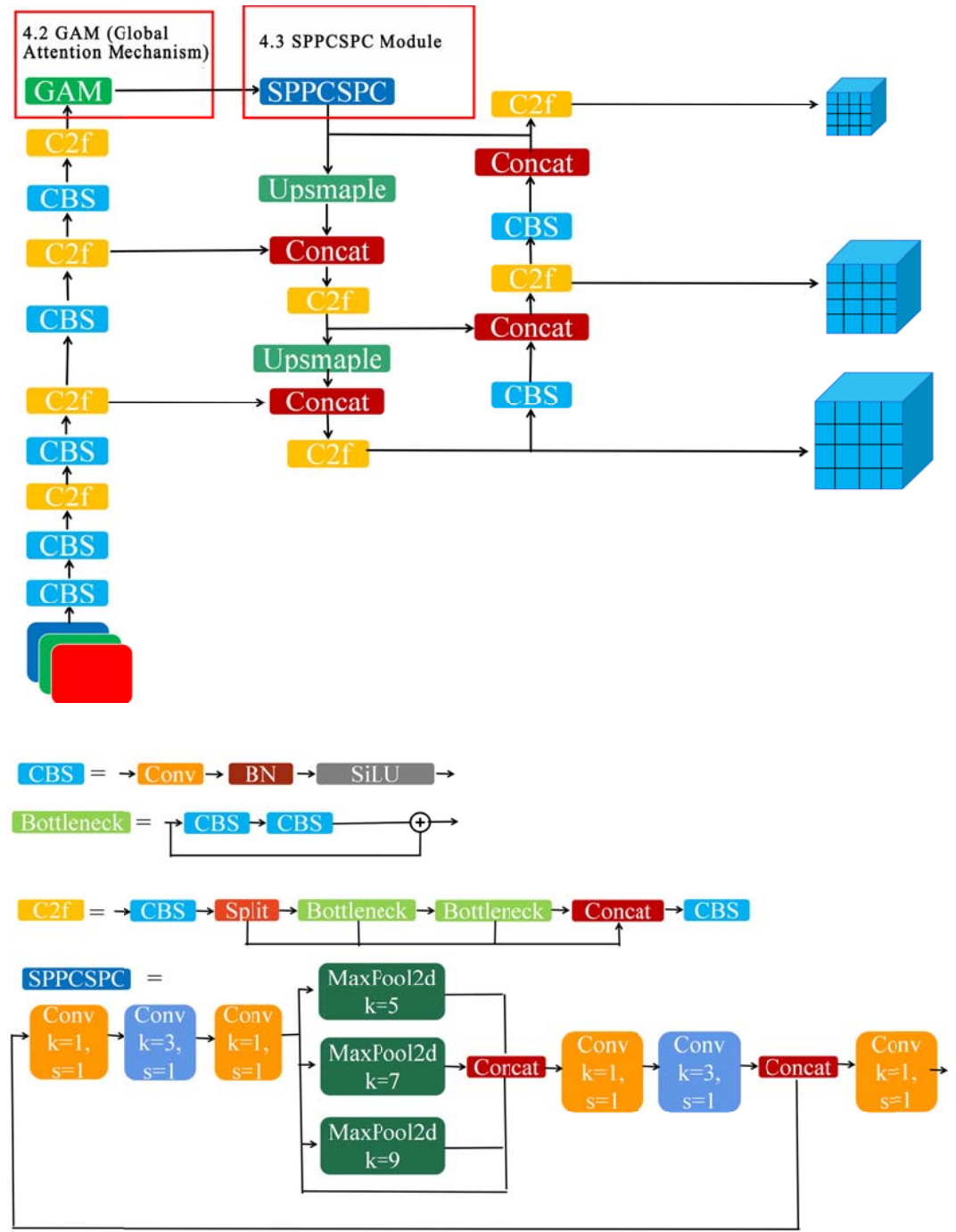

**Figure 3.** Improved YOLOv8m Model for Foreign-Object Detection in Transmission Lines.

### 4.2. GAM (Global Attention Mechanism)

Specifically addressing the issue of occlusion in detecting foreign objects on power transmission lines, we integrate the GAM attention mechanism into the backbone to enhance the model's focus on the foreign object.

The Global Attention Mechanism (GAM) represents a technique that enriches inter-dimensional interactions across channel and spatial dimensions by preserving channel and spatial information. This augmentation is pivotal in obtaining critical target information.

GAM adopts the channel-spatial sequence of CBAM [45] and entails a redesign of both the channel-attention submodule and the spatial-attention submodule, depicted in

Figure 4. In Formula (1), $M_C$ represents the handling of the input vector $F_1$ by the channel-attention submodule, where $\otimes$ is the tensor product of the processed vector and the input vector $F_1$. In Formula (2), $M_S$ signifies the spatial-attention submodule's processing of the vector $F_2$, where $\otimes$ is the tensor product of the processed vector and the input vector $F_2$.

$$F_2 = M_C(F_1) \otimes F_1 \qquad (1)$$

$$F_3 = M_S(F_2) \otimes F_2 \qquad (2)$$

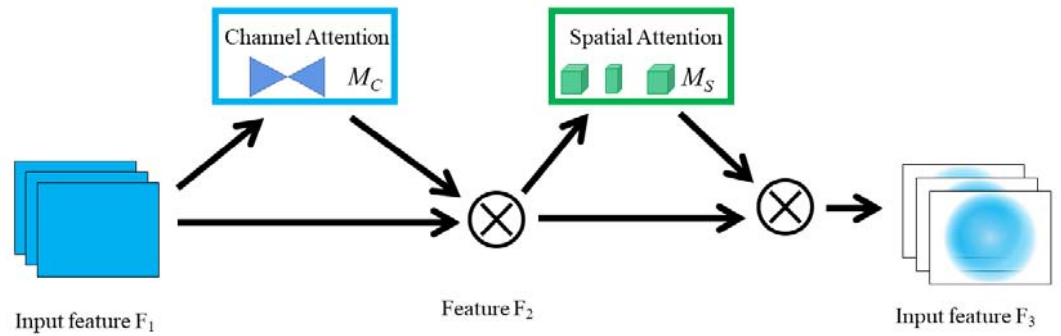

**Figure 4.** GAM structure diagram.

In Figure 5, we delineate the specific processing procedure of the channel-attention submodule depicted in Figure 4. The channel-attention submodule employs a 3D permutation (dimensional transformation) to retain three-dimensional information. Subsequently, a two-layer Multi-Layer Perceptron (MLP) enhances inter-dimensional dependency relationships spanning channel and spatial dimensions. An MLP structure is an encoder–decoder architecture with a reducing factor of r.

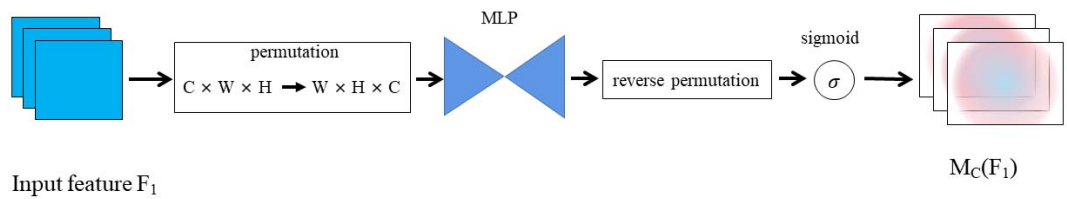

**Figure 5.** Channel-attention submodule diagram.

In Figure 6, we delineate the specific processing procedure of the spatial-attention submodule depicted in Figure 4. The spatial-attention submodule employs two convolutions to fuse spatial information, to heighten focus on spatial information. Moreover, a reducing factor r is used within the channel-attention submodule to augment attention towards spatial features, as depicted in Figure 6.

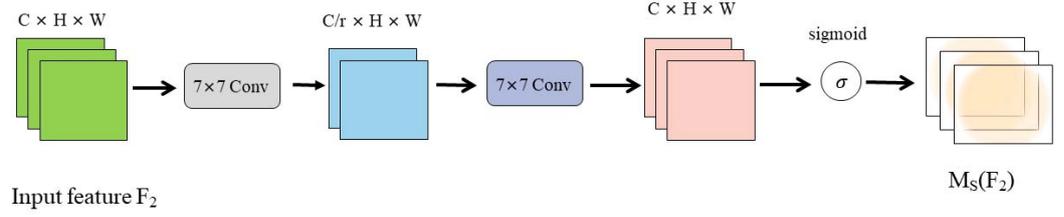

**Figure 6.** Spatial-attention submodule diagram.

### 4.3. SPPCSPC Module

In response to significant scale changes, multiple types and complex backgrounds of foreign objects, we have replaced the SPPF module with the SPPCSPC module, to augment the model's ability to extract multi-scale features. The SPPCSPC module uses a multi-scale pyramid pooling (SPP) structure and conventional convolutional operations. This structure amplifies the model's receptive field and elevates its capacity to extract features across multiple scales, enhancing overall detection capabilities.

The SPPCSPC module, shown in Figure 7, segregates features into two segments. One segment undergoes processing utilizing the SPP structure, employing four max-pooling operations at distinct scales, to capture diverse receptive fields. Consequently, this step amalgamates spatial feature information of varying dimensions and generates fixed-size feature vectors. In parallel, the other segment undergoes standard convolutional operations. Finally, the features obtained from these two segments are concatenated to form the comprehensive feature vector.

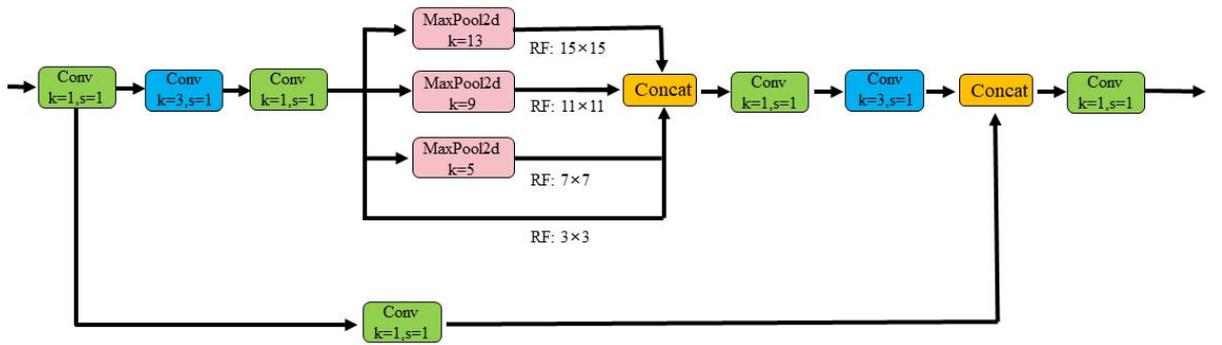

**Figure 7.** SPPCSPC Model diagram.

### 4.4. Focal-EIoU Loss Function

To address the imbalance between high- and low-quality samples (the intersection size of different regions on the union of predicted-truth boxes and ground-truth boxes), we add a Focal-EIou loss function. The Focal-EIoU loss function enhances the optimization contributions for high-quality samples with lower quantities when regressing the predicted bounding boxes towards the ground-truth boxes, while simultaneously diminishing the optimization contributions for low-quality samples with higher quantities in this regression process. The Focal-EIoU loss function can effectively address the imbalance between high- and low-quality samples, expediting model convergence and augmenting detection precision.

The EIoU loss function, shown in Figure 8, is represented by Equation (3). The EIoU is a metric for evaluating the dissimilarity between predicted boxes and ground-truth boxes. The Focal-EIoU loss function utilizes this metric to guide the training process, emphasizing high-quality samples for more effective model learning and convergence.

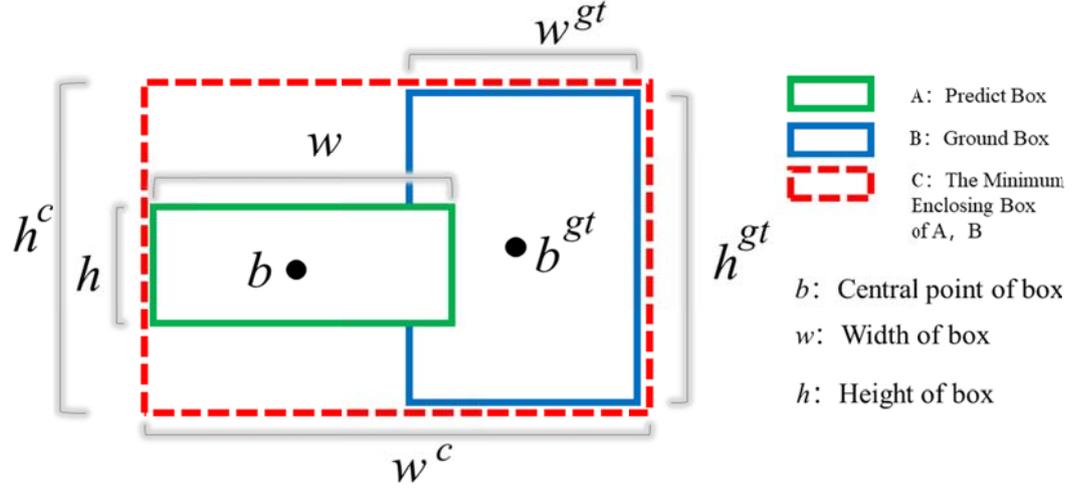

**Figure 8.** EIoU loss function diagram.

$$L_{EIoU} = L_{IoU} + L_{dis} + L_{asp} = 1 - IoU(A,B) + \frac{\rho^2(b,b^{gt})}{(h^c)^2+(w^c)^2} + \frac{\rho^2(h,h^{gt})}{h^c} + \frac{\rho^2(w,w^{gt})}{w^c} \qquad (3)$$

As illustrated in Equation (3) and Figure 8 above, where $1-IoU(A,B)$ represents the loss function for the predicted box and the ground-truth box, $\rho(\cdot)$ signifies the Euclidean distance, $b$ and $b^{gt}$ denote the centers of the predicted and ground-truth boxes, respectively, $h^c$ and $w^c$ denote the height and width of the bounding box, $h$ and $h^{gt}$ denote the height of the predicted and ground-truth boxes, and $w$ and $w^{gt}$ denote the widths of the predicted and ground-truth boxes, respectively.

The EIoU loss function consists of three components: the Intersection over Union (IoU) loss between the predicted and ground-truth boxes, the center-distance loss between the predicted and ground-truth boxes, and the edge-length loss between the predicted and ground-truth boxes. The EIoU loss function directly minimizes the differences in width and height between the predicted and ground-truth bounding boxes, expediting model convergence and enhancing localization accuracy.

After adding Focal to EIoU, EIoU focuses more on high-quality samples (large Intersection over Union between predicted and ground-truth boxes). The comprehensive formula for Focal-EIoU is given by Equation (4).

$$L_{Focal-EIoU} = IoU^{\gamma} L_{EIoU} \qquad (4)$$

where $IoU$ represents the Intersection over Union for predicted box A and ground-truth box B ( $IoU = (A \cap B)/(A \cup B)$ ), $\gamma$ is a parameter controlling the degree of suppression for low-quality samples (small Intersection over Union between predicted and ground-truth boxes), and $L_{EIoU}$ represents the EIoU loss function.

## 5. Experiment and Analysis

### 5.1. Experimental Setup

The hardware configuration for our experiment included an Intel(R) Core(TM) i7–10700K CPU, 32 GB of RAM, and an NVIDIA GeForce RTX 2080 Ti GPU. The software environment comprised Windows 10, PyTorch 1.11.0, CUDA 11.3, and PyCharm Community Edition 2021.3.

The momentum was initialized to 0.937, the batch size was 24, the weight decay rate was 0.0005, the initial learning rate was 0.01, and the training epoch was 100. Given the varying sizes of the image samples, all input images were transformed to 640 × 640 pixels.

*5.2. Evaluation Metrics*

In object detection, Intersection over Union (IoU) represents the area ratio of the intersection to the union between the predicted box and the ground-truth box. TP (True Positive) denotes the sample count where the real category is positive and the model's prediction is also positive; TN (True Negative) indicates the sample count where the real category is negative, and the model predicts them as negative; FP (False Positive) is the sample count where the real category is negative, but the model predicts them as positive; and FN (False Negative) represents the sample count where the real category is positive, but the model predicts them as negative.

Our ablation experiments used seven metrics: mAP_0.5, mAP_0.5:0.95, precision, recall, parameters, GFloats, and speed. Our comparative experiments used five metrics: mAP_0.5, mAP_0.5:0.95, precision, recall, and speed. The definitions of these metrics are as follows:

1. Precision: the proportion of predicted positive samples that are actually positive (Precision = TP/(TP + FP)).
2. Recall: the proportion of actual positive samples that are predicted as positive (Recall = TP/(TP + FN)).
3. mAP_0.5: precision is calculated for each class when the IoU threshold is 0.5, and the precision of each class is averaged to obtain mAP_0.5.
4. mAP_0.5:0.95: average mAP across different IoU thresholds from 0.5 to 0.95 (with a step of 0.05).
5. Parameter: the model's parameter count.
6. GFLOPs: model computational complexity. One GFLOPs means that the model requires billions of floating-point operations.
7. Speed: the model's detection speed. It indicates the milliseconds required for detecting a single image.

*5.3. Experimental Results and Analysis*

5.3.1. Model Training Comparison

In this experiment, the training epoch was 100. Figure 9 illustrates the comparison of mAP_0.5 during training, Figure 10 presents the comparison of mAP_0.5:0.95 during training, Figure 11 exhibits the comparison of precision during training, Figure 12 showcases the comparison of recall during training, and Figure 13 displays the comparison of cls_loss during training. From these figures, it is evident that our model outperforms the original model across various metrics.

The four curves at the beginning of "Original" are trained using the original dataset, and their models are the original model, incorporating the GAM attention mechanism, continuing to add the SPPCSCP module, and continuing to add the Focal -EIoU loss function, respectively. The other two curves, starting with "dataAugmentation", are trained using the augmented dataset, and their models are the original model and our improved model, respectively.

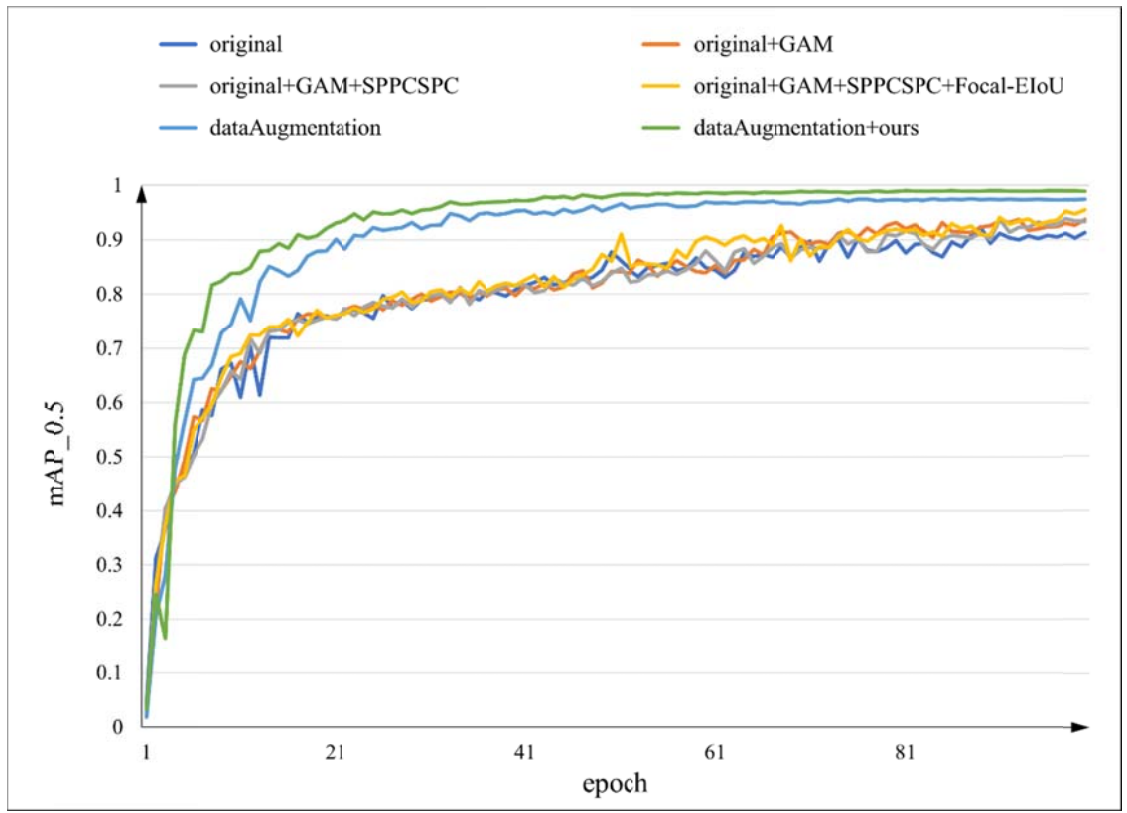

**Figure 9.** Comparison of mAP_0.5 during training.

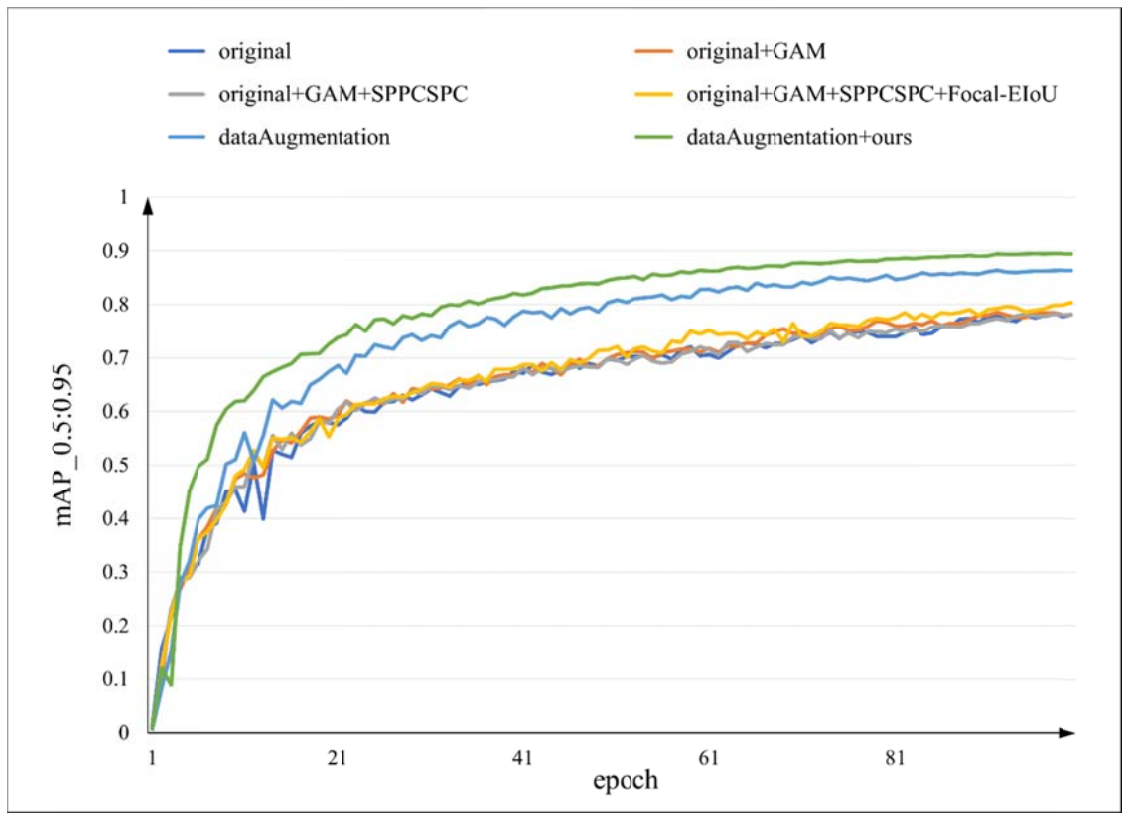

**Figure 10.** Comparison of mAP_0.5:0.95 during training.

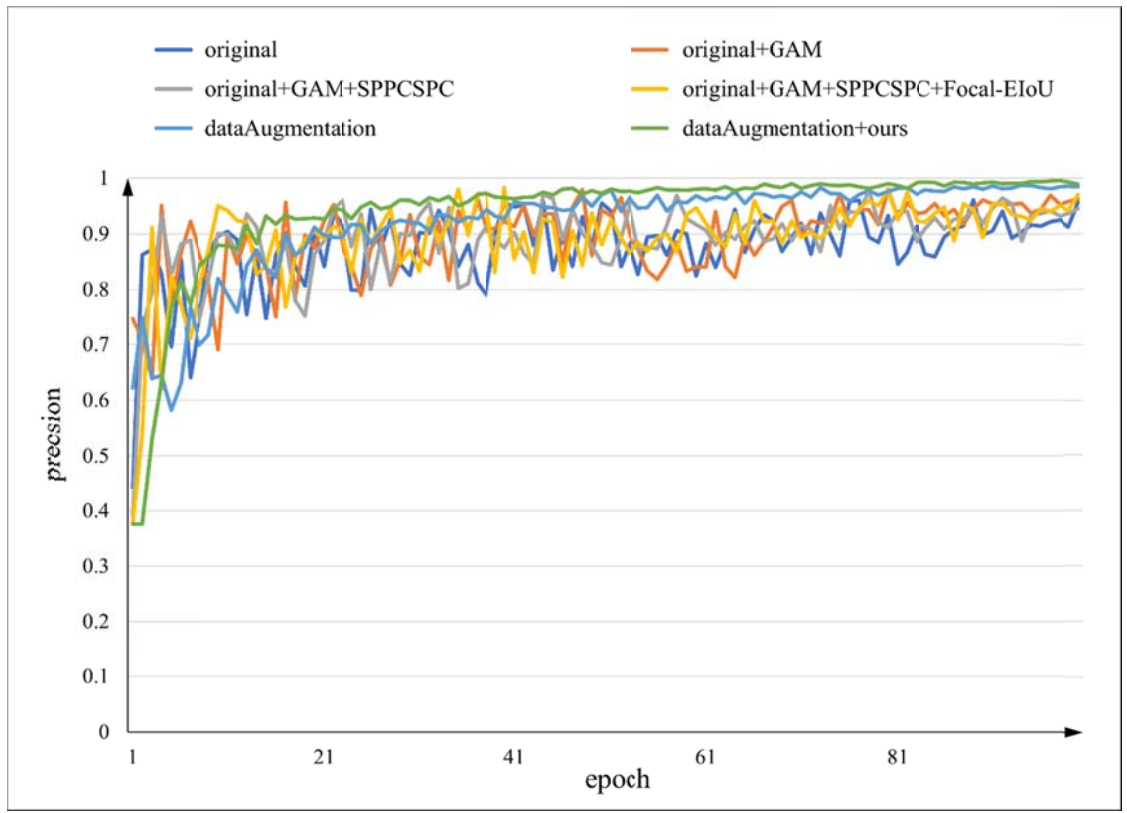

**Figure 11.** Comparison of precision during training.

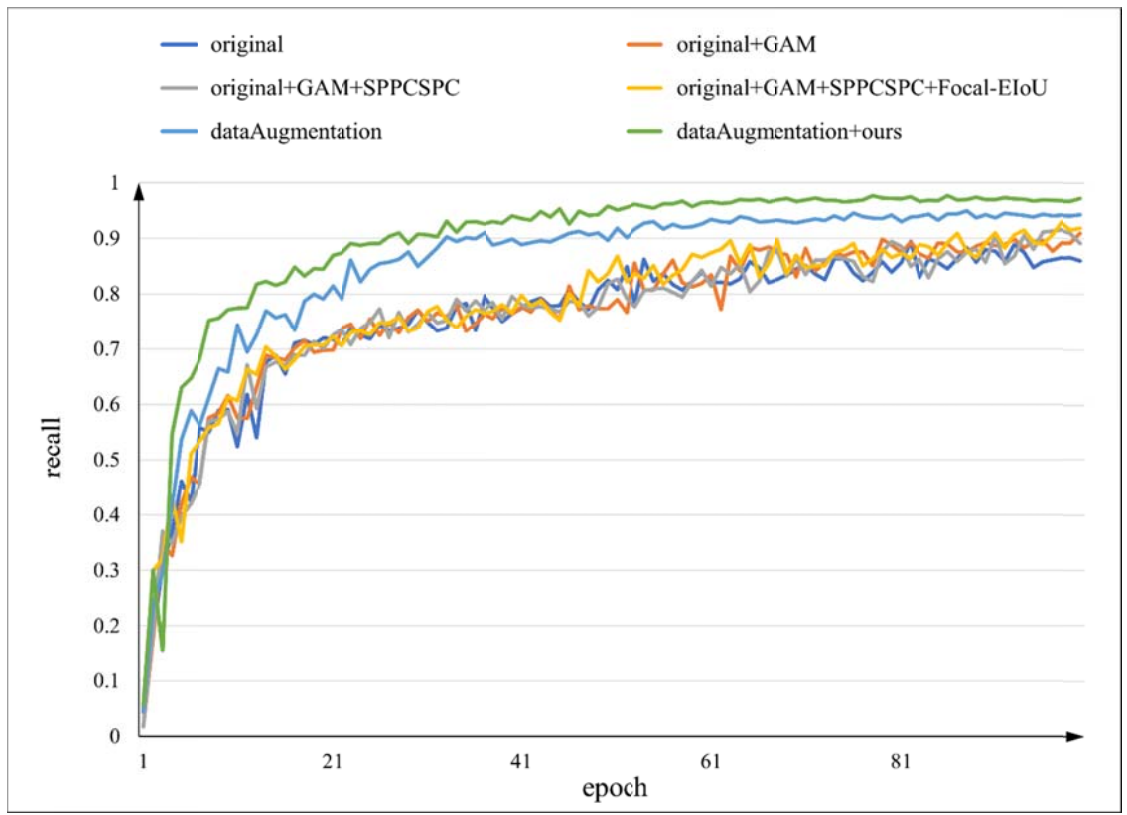

**Figure 12.** Comparison of recall during training.

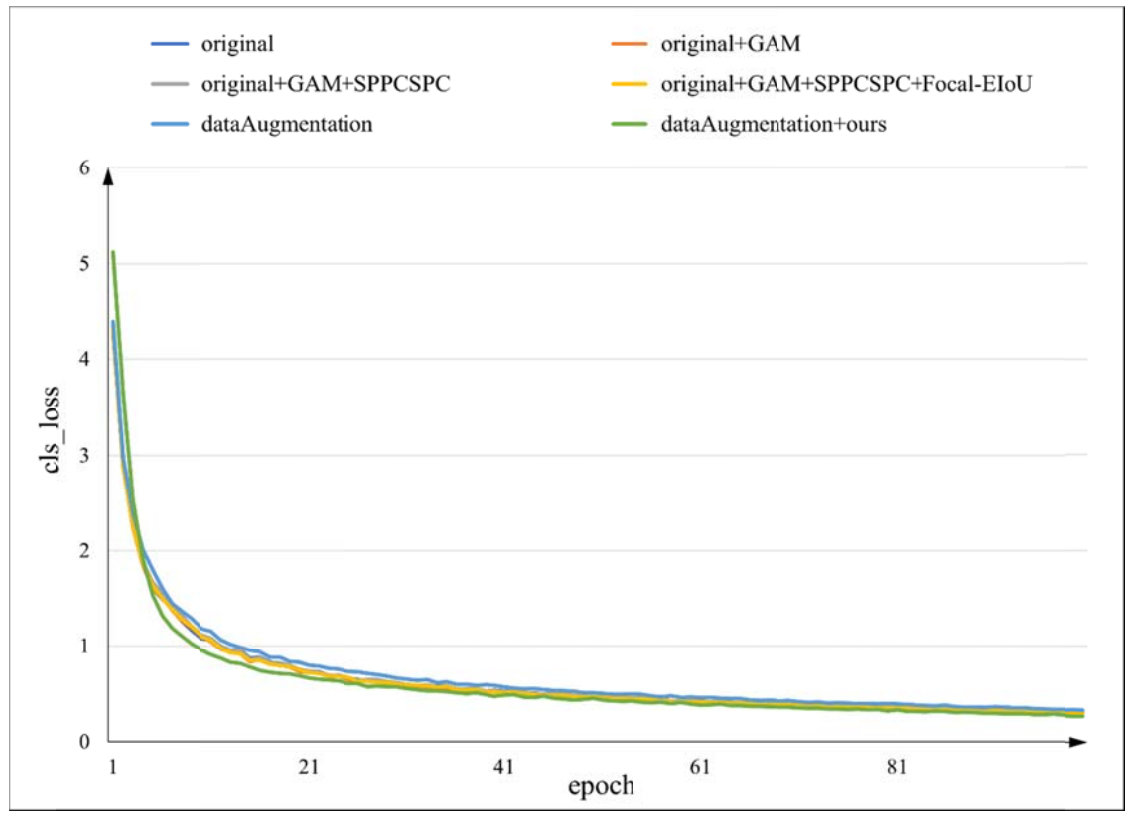

**Figure 13.** Comparison of cls_loss during training.

5.3.2. Attention Visualization Comparison

Figure 14 compares attention visualization results between the YOLOv8m original model and the addition of the GAM attention mechanism. Incorporating the GAM attention mechanism into our model enhances the focus on the objects and diminishes attention toward the background. The GAM can improve detection confidence and precision.

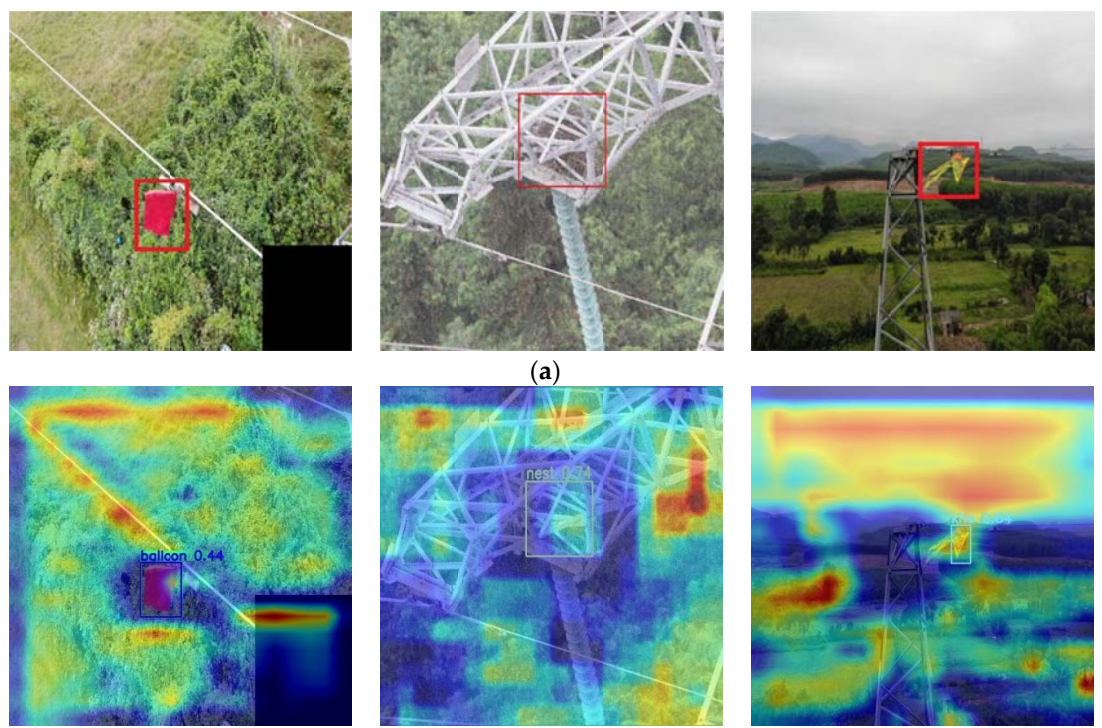

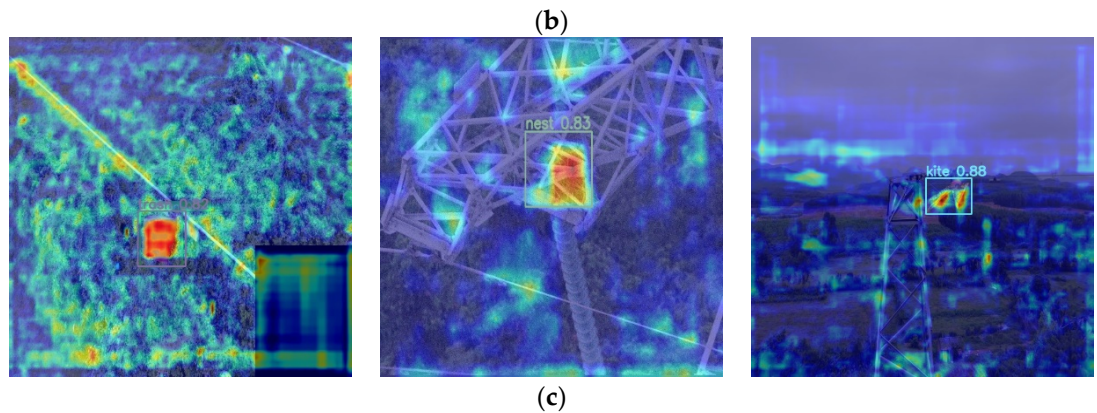

(c)

**Figure 14.** Attention visualization comparison. (a) original images. (b) Attention visualization using YOLOv8m. (c) Attention visualization using GAM.

### 5.3.3. Ablation Experiments

To demonstrate the effectiveness of the improvement steps, we conducted ablation experiments, and the results are presented in Table 2.

As shown in Table 2, different improved steps can improve mAP_0.5, mAP_0.5:0.95, and recall. After simultaneously incorporating the GAM attention mechanism, SPPCSPC, and Focal-EIoU loss function, the mAP_0.5 increased by 2.7%, mAP_0.5:0.95 improved by 4%, and recall increased by 6%. Remarkably, the recall showed the most significant improvement. In practical applications, secondary manual detection is often performed on detected images. A higher recall can effectively reduce the miss-detection rate, subsequently reducing the workload for detection personnel.

**Table 2.** Results of Ablation Experiment.

| GAM | SPPCSPC | Focal-EIoU | mAP_0.5 (%) | mAP_0.5:0.95 (%) | Precision (%) | Recall (%) | Parameters | GFLPOS | Speed (ms) |
|---|---|---|---|---|---|---|---|---|---|
| | | | 92.8 | 76.4 | 98.4 | 85.9 | 25.9 M | 79.1 | 32.3 |
| √ | | | 93.1 | 78.5 | 95.4 | 88.9 | 42.5 M | 85.6 | 34.5 |
| | √ | | 93.5 | 73.71 | 93.96 | 89.22 | 34 M | 92.4 | 33.7 |
| | | √ | 94.14 | 75.12 | 94.6 | 89.9 | 25.9 M | 79.1 | 32.4 |
| √ | √ | | 93.9 | 78.3 | 93.2 | 91.5 | 50.6 M | 98.9 | 35.9 |
| | √ | √ | 93.75 | 82.1 | 98 | 90.86 | 34 M | 92.4 | 33.9 |
| √ | | √ | 94.89 | 84.5 | 97.03 | 91.09 | 42.5 M | 85.6 | 34.8 |
| √ | √ | √ | 95.5 | 80.4 | 96.7 | 91.9 | 50.6 M | 98.9 | 35.9 |

### 5.3.4. Comparative Experiments with Other Models

To validate the advancement of our improved model, we compared it with mainstream object-detection models, using the same dataset and dataset partition. The comparative experimental results are presented in Table 3.

**Table 3.** Comparative Experiments with Other Models.

| Model | mAP_0.5 (%) | mAP_0.5:0.95 (%) | Precision (%) | Recall (%) | Speed (ms) |
|---|---|---|---|---|---|
| Faster RCNN | 86.7 | 64.9 | 81.9 | 83.1 | 80.9 |
| Sparse RCNN | 90 | 66.4 | 93.8 | 81.1 | 79.1 |
| Cascade RCNN | 86.8 | 65.3 | 89.1 | 77.7 | 86.7 |
| YOLOv5m | 91.9 | 68.3 | 95 | 86.2 | 50.5 |
| YOLOv6m | 88.67 | 71.8 | 91.8 | 86.5 | 45.7 |
| YOLOv7 | 92 | 75.9 | 98.2 | 85.5 | 41.5 |
| YOLOv8m | 92.8 | 76.4 | 98.4 | 85.9 | 32.3 |
| Ours | 95.5 | 80.4 | 96.7 | 91.9 | 35.9 |

As depicted in Table 3, one-stage methods generally perform better than two-stage models. Regarding mAP_0.5, mAP_0.5:0.95, and recall metrics, our model stands out, achieving the highest values at 95.5%, 80.4%, and 91.9%, respectively. While not the fastest in terms of speed, it ranks second, and in terms of precision, it holds the third position. Our model demonstrates superior performance compared to other detection models, notably achieving the highest values for mAP_0.5, mAP_0.5:0.95, and recall.

### 5.3.5. Foreign-Object Detection Result

Figure 15 illustrates the detection results of the original YOLOv8m model, where (a), (b), (c), (d), (e), and (f) correspond to trash, twig, nest, kite, bird and balloon, respectively. Figure 16 presents the detection results of the improved YOLOv8m model, with each subfigure corresponding to the respective categories shown in Figure 15.

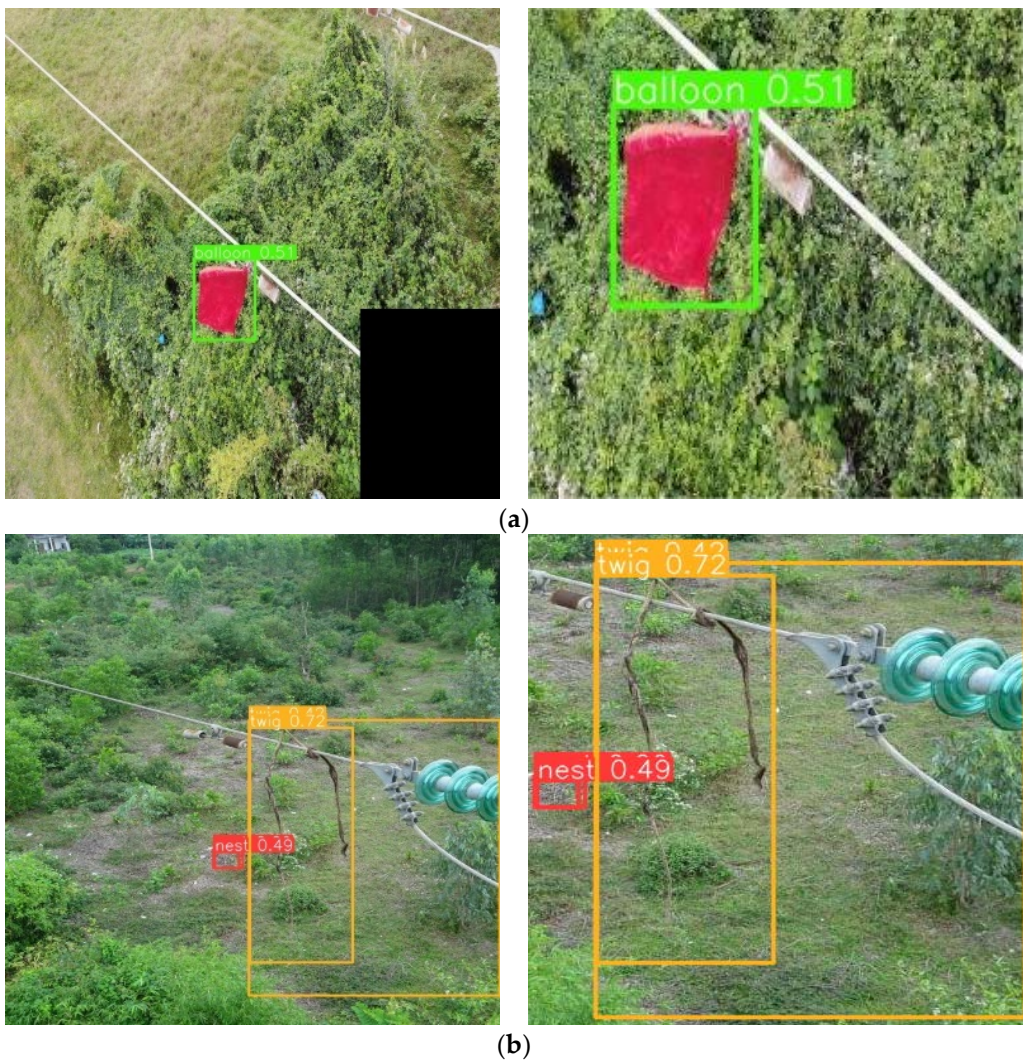

(a)

(b)

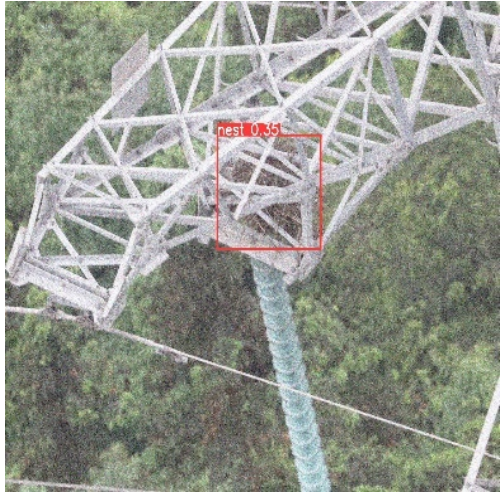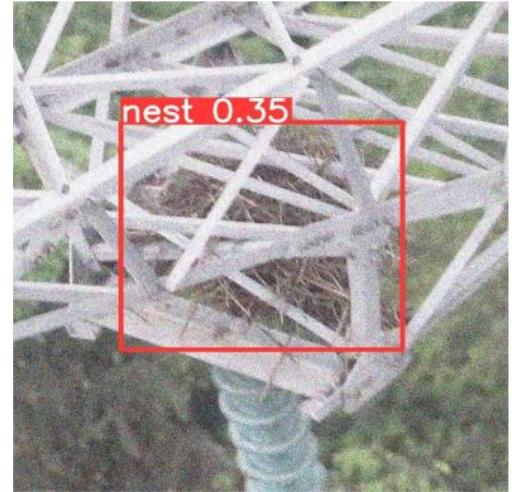

(c)

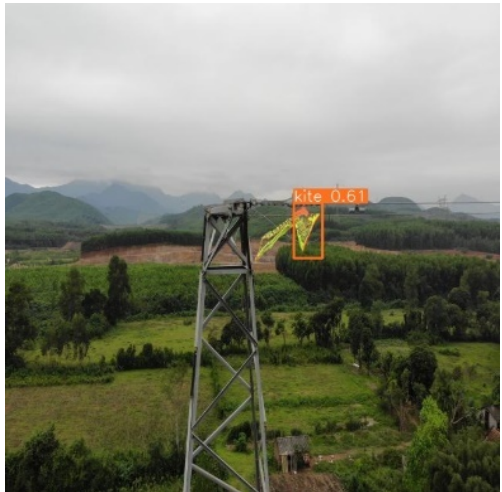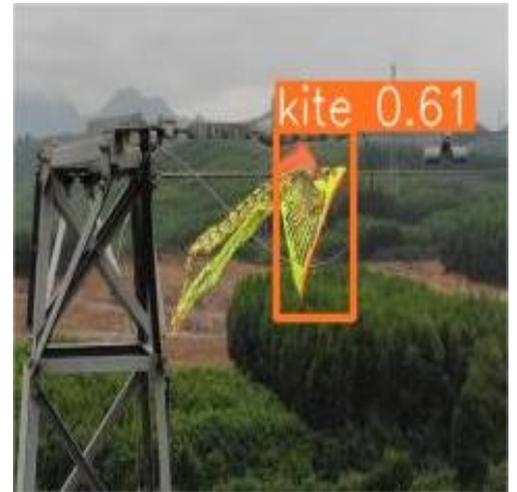

(d)

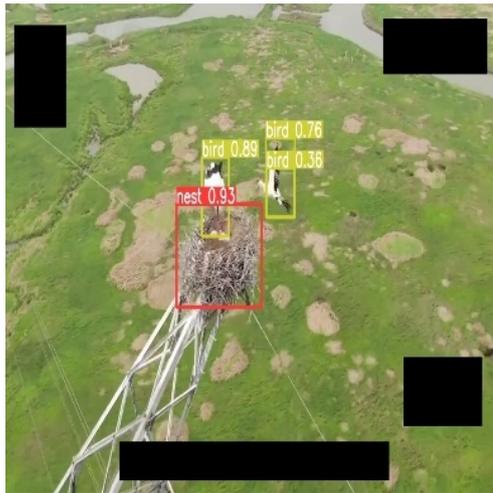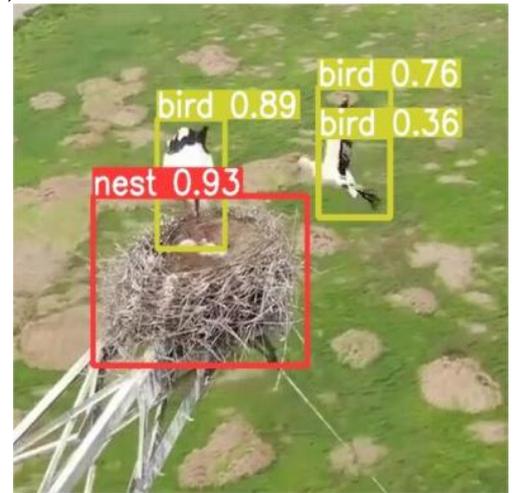

(e)

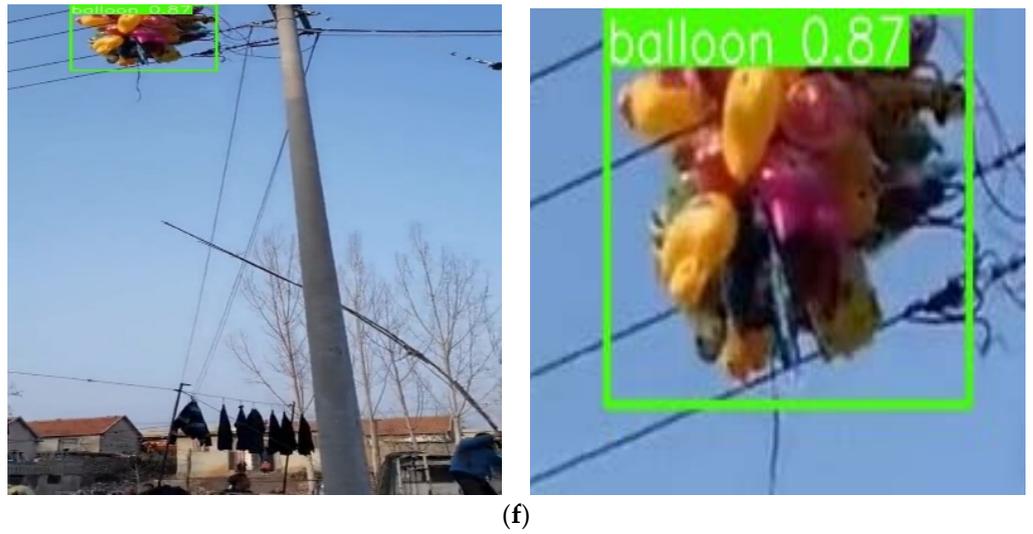

(**f**)

**Figure 15.** YOLOv8m original-model detection results and enlarged images (left column: original image of detection results, right column: enlarged image). (**a**) trash. (**b**) twig. (**c**) nest. (**d**) kite. (**e**) bird. (**f**) balloon.

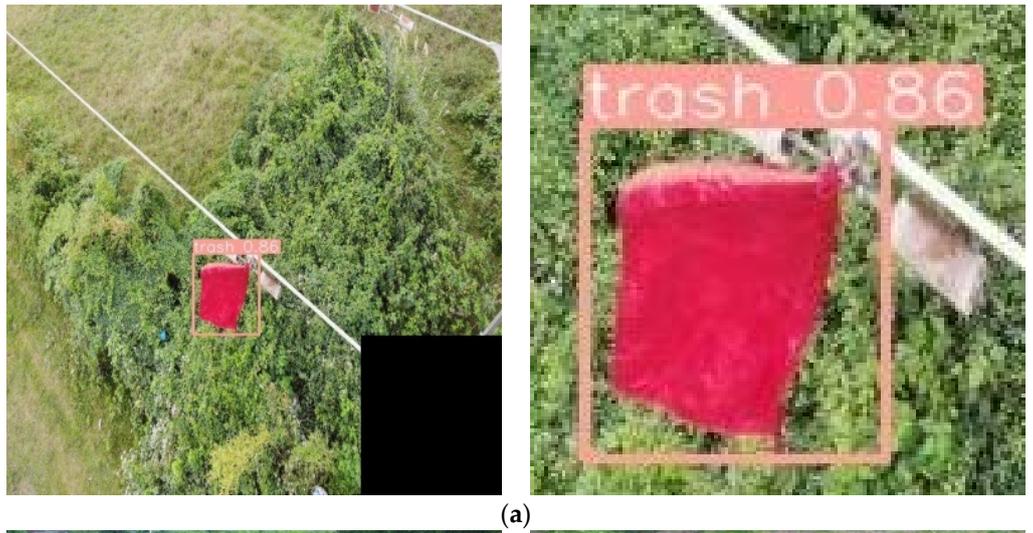

(**a**)

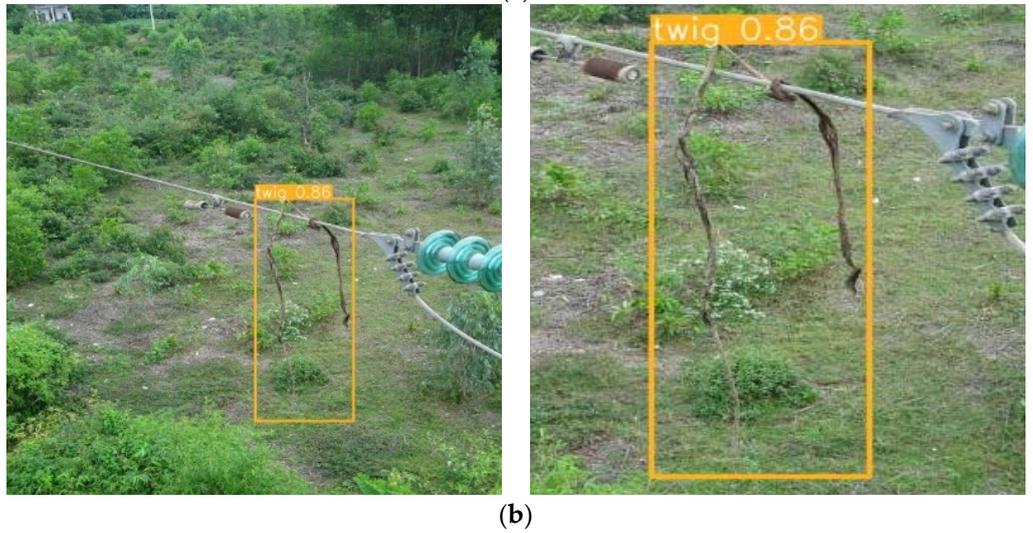

(**b**)

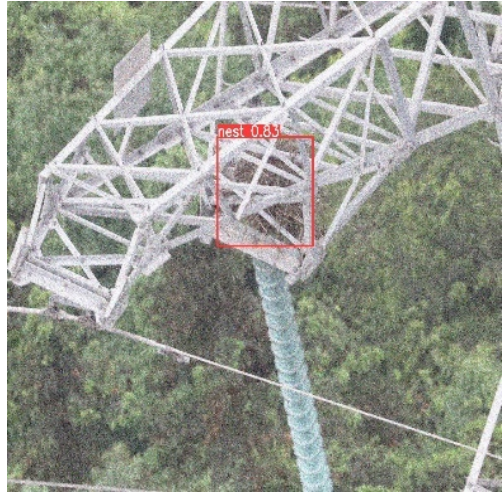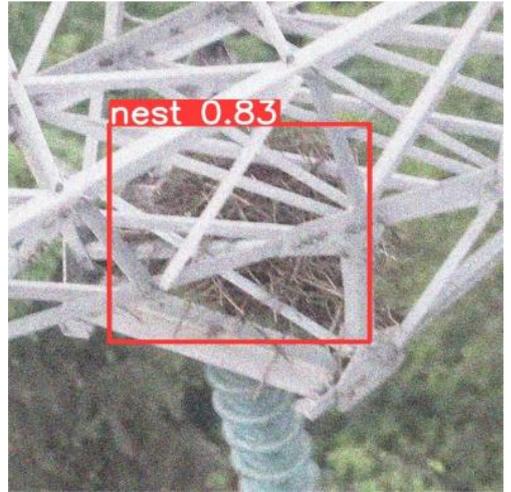

(**c**)

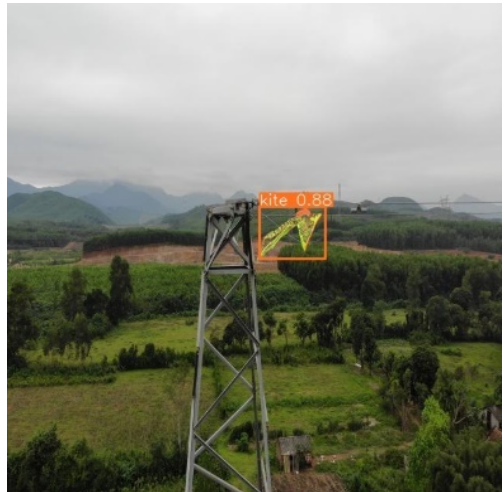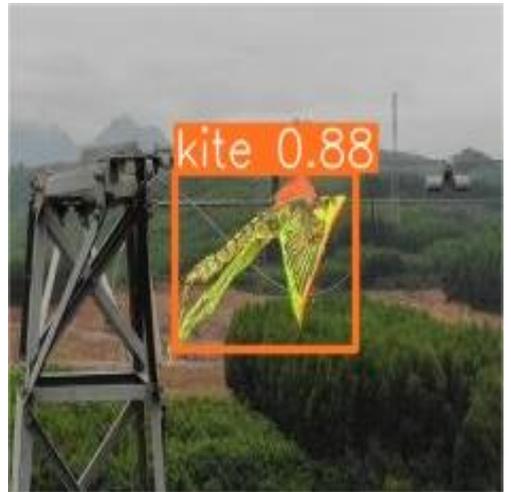

(**d**)

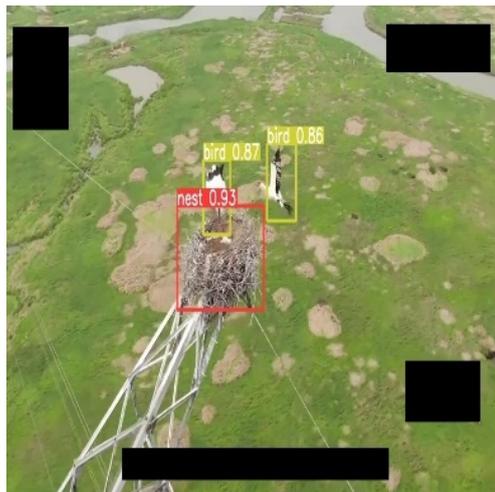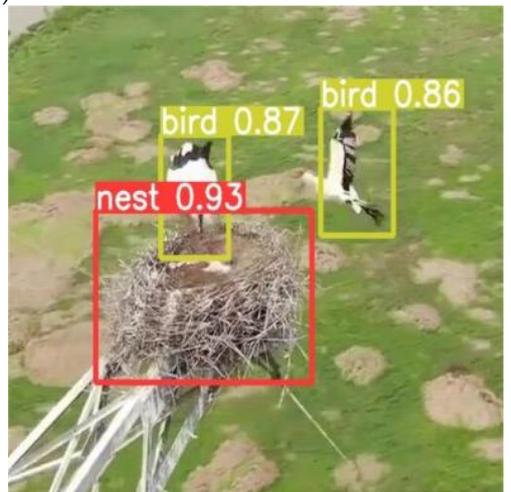

(**e**)

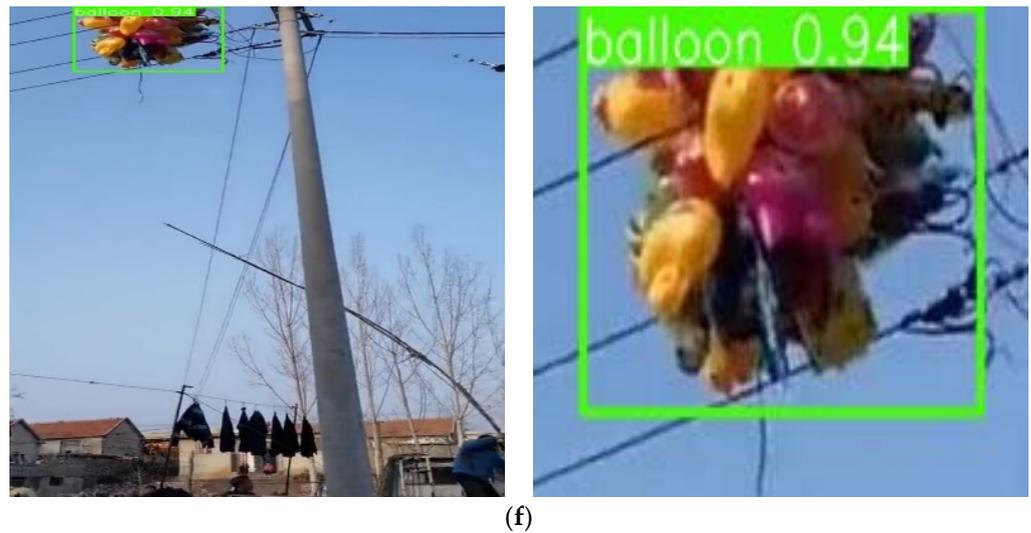

(f)

**Figure 16.** Improved YOLOv8m-model detection results and enlarged images. (left column: original image of detection results, right column: enlarged image). (a) trash. (b) twig. (c) nest. (d) kite. (e) bird. (f) balloon.

In Figures 15 and 16, we sequentially compare the detection results for each subfigure. In Subfigure15(a) and 16(a), our improved model enhances the confidence in detecting trash targets. In Subfigure 15(b) and 16(b), our improved model eliminates duplicate predictions of different parts of the same twig and false detections of a nest, while also achieving a higher confidence in twig detection. In Subfigure 15(c) and 16(c), our improved model exhibits higher confidence in detecting an obscured nest. In Subfigure 15(d) and 16(d), our improved model avoids incomplete detection of kite targets and enhances confidence in detecting kite targets. In Subfigure 15(e) and 16(e), our improved model eliminates duplicate predictions of different parts of the same bird and improves confidence in bird target detection. Finally, in Subfigure15(f) and 16(f), our improved model elevates the confidence in detecting balloon targets.

### 5.3.6. Detection Results for Different Classes

Table 4 presents a comparative analysis of different models in detecting various foreign objects.

**Table 4.** Detection Results of Various Foreign Objects.

| Model | Foreign Object | Labels | mAP_0.5(%) | mAP_0.5:0.95 (%) | Precision (%) | Recall (%) |
|-------|----------------|--------|-----------|-----------------|---------------|-----------|
| YOLOv8m | all | 1008 | 92.8 | 76.4 | 98.4 | 85.9 |
| | nest | 514 | 98.7 | 85.9 | 99.4 | 93.3 |
| | trash | 134 | 98.4 | 84.9 | 100 | 92.5 |
| | kite | 45 | 86.3 | 70.1 | 99.2 | 71.1 |
| | twig | 14 | 76.1 | 37.9 | 93.2 | 64.3 |
| | bird | 242 | 99.5 | 89.6 | 99.6 | 99.2 |
| | balloon | 59 | 97.4 | 89.9 | 99.3 | 94.9 |
| Ours | all | 1008 | 95.5 | 80.4 | 96.7 | 91.9 |
| | nest | 514 | 98.7 | 87.3 | 98.6 | 96.3 |
| | trash | 134 | 98.7 | 88.4 | 98.2 | 94.8 |
| | kite | 45 | 87.9 | 75.8 | 87.3 | 80 |
| | twig | 14 | 90.4 | 48.3 | 97.1 | 85.7 |
| | bird | 242 | 99.5 | 90.1 | 99.4 | 99.6 |
| | balloon | 59 | 97.6 | 92.8 | 99.4 | 94.9 |

| | | | | | |
|---|---|---|---|---|---|
| | all | 2798 | 97.4 | 86.4 | 98.6 | 94.1 |
| | nest | 1004 | 99.1 | 88.4 | 99.8 | 98 |
| | trash | 255 | 93.9 | 86.1 | 97.8 | 90.5 |
| Data Augmenta-tion | kite | 457 | 96.1 | 81.5 | 97.1 | 90.3 |
| | twig | 246 | 99.3 | 82.3 | 98.4 | 97.1 |
| | bird | 487 | 99 | 90.4 | 98.9 | 97.2 |
| | balloon | 349 | 96.9 | 89.7 | 99.4 | 91.7 |
| | all | 2798 | 99 | 89.5 | 99.6 | 96.8 |
| | nest | 1004 | 99.4 | 89.8 | 99.5 | 97.8 |
| Data Augmenta-tion + Ours | trash | 255 | 98 | 91 | 99.7 | 94.5 |
| | kite | 457 | 98.1 | 87.1 | 98.8 | 92.3 |
| | twig | 246 | 99.5 | 84.7 | 100 | 97.5 |
| | bird | 487 | 99.4 | 91.8 | 99.4 | 99 |
| | balloon | 349 | 99.5 | 93 | 100 | 99.4 |

From Table 4, it is evident that before data augmentation, our improved model (denoted as 'Ours') shows an overall increase in mean precision across all categories in terms of mAP_0.5, mAP_0.5:0.95, and recall when compared to the original YOLOv8m model. However, there is a decrease in precision. Specifically, for mAP_0.5, the 'twig' category exhibits the most significant improvement, with other categories also showing improvements or remaining relatively stable. Similarly, in mAP_0.5:0.95, 'twig' demonstrates the most substantial advancement, while the other categories also display enhancements or stability. Regarding precision, except for 'twig' and 'balloon', which experience improvements, the precision of other categories decreases. For recall, all categories witness an increase or remain stable.

Following data augmentation, a comparison between our improved model ('Ours') and the original YOLOv8m model reveals an overall increase in mean precision across all categories concerning mAP_0.5, mAP_0.5:0.95, precision, and recall. In mAP_0.5, the precision of each category rises. In mAP_0.5:0.95, the precision of each category shows improvement. Regarding precision, 'nest' experiences a decline, but the precision of other categories improves. In recall, 'nest' shows a decline, but the precision of other categories sees an increase.

## 6. Conclusions

This paper proposes a foreign-object-detection model for transmission lines based on improved YOLOv8m. Our model autonomously detects anomalies in aerial images of power transmission lines, reducing the workload associated with anomaly detection and enhancing inspection efficiency.

Addressing the challenges of foreign-object detection along power transmission lines, we use three key improvements: firstly, to mitigate issues related to occlusion, a Global Attention Mechanism (GAM) was introduced, enhancing focus on obscured targets. Secondly, in response to the diversity of foreign-object categories and complex backgrounds, the SPPCSPC module was integrated, boosting the model's multi-scale feature-extraction capabilities. Thirdly, the Focal-EIoU loss function was introduced to address imbalances between high- and low-quality samples, expediting model convergence and augmenting detection accuracy. The refined YOLOv8m model exhibited significant enhancements, elevating mAP_0.5 from 92.8% to 95.5%, mAP_0.5:0.95 from 76.4% to 80.4%, and recall from 85.9% to 91.9%.

No publicly available datasets for power transmission line foreign-object detection have been provided in previously published literature, thus hindering experimentation with our enhanced model on alternative datasets. The dataset utilized in this study originates from the Electric Power Research Institute, Yunnan Power Grid Co., Ltd., featuring geolocation data in certain images. Upon removing location information and ob-taining approval from the aforementioned organization, we plan to release this dataset publicly, facilitating future research in related domains.

Although the upgraded model demonstrates the potential to be applied in practical detection scenarios for foreign objects on power transmission lines, its detection speed is presently insufficient for direct application on terminal CPU devices. Subsequent work will focus on adopting lighter models, such as YOLOv8n, to enhance model detection speed while maintaining detection accuracy.

**Author Contributions:** Conceptualization, Z.W.; methodology, Z.W.; software, Z.W.; validation, Z.W.; formal analysis, Z.W., G.Y., H.Z., Y.M. (Yi Ma) and Y.M. (Yutang Ma); investigation, Z.W., G.Y., H.Z., Y.M. (Yi Ma) and Y.M. (Yutang Ma); resources, Z.W. and G.Y.; data curation, Z.W.; writing—original draft preparation, Z.W. and G.Y.; writing—review and editing, Z.W. and G.Y.; funding acquisition, G.Y., H.Z., Y.M. (Yi Ma) and Y.M. (Yutang Ma) All authors have read and agreed to the published version of the manuscript.

**Data Availability Statement:** The data presented in this study are available on request from the corresponding author. The data are not publicly available due to [The data is provided by Yunnan Power Grid Co., Ltd. and can only be released after being approved by Yunnan Power Grid Co., Ltd..]

**Acknowledgments:** Thank you to Yunnan Power Grid Co., Ltd. for providing the foreign object dataset.

**Conflicts of Interest:** Author Zhenyue Wang, Yi Ma and Yutang Ma are employed by the Yunnan Power Grid Co., Ltd. The remaining authors declare that the research was conducted in the absence of any commercial or financial relationships that could be construed as a potential conflict of interest.